\documentclass[10pt,twocolumn,letterpaper]{article}

\usepackage[pagenumbers]{cvpr} % To force page numbers, e.g. for an arXiv version

\usepackage{amsmath,amssymb,amsfonts}
\usepackage{algorithm}
\usepackage{array}
\usepackage{subcaption}
\usepackage{tikz}
\usepackage{textcomp}
\usepackage{stfloats}
\usepackage{url}
\usepackage{bm}
\usepackage{listings}
\usepackage{verbatim}
\usepackage{graphicx}
\usepackage{subfiles}
\usepackage{amsfonts}
\usepackage{booktabs}
\usepackage{multirow}
\usepackage{pifont}
\usepackage{algpseudocode}
\usepackage{siunitx}
\definecolor{cvprblue}{rgb}{0.21,0.49,0.74}
\usepackage[pagebackref,breaklinks,colorlinks,allcolors=cvprblue]{hyperref}

\def\paperID{*****} % *** Enter the Paper ID here
\def\confName{CVPR}
\def\confYear{2025}

\title{Taming the Randomness: Towards Label-Preserving Cropping in Contrastive Learning}

\author{
Mohamed Hassan$^{\dagger}$, Mohammad Wasil$^{*}$, Sebastian Houben$^{*}$\\
$^{\dagger,*}$Hochschule Bonn-Rhein-Sieg\\
Grantham-Allee 20, 53757 Sankt Augustin, Germany\\
{\tt\small $^\dagger$mohamed.hassan@smail.inf.h-brs.de}\\
{\tt\small $^*$\{firstname.lastname\}@h-brs.de}
}
\begin{document}
\maketitle
\begin{abstract}
Contrastive learning (CL) approaches have gained great recognition as a very successful subset of self-supervised learning (SSL) methods. SSL enables learning from unlabeled data, a crucial step in the advancement of deep learning, particularly in computer vision (CV), given the plethora of unlabeled image data. CL works by comparing different random augmentations (e.g., different crops) of the same image, thus achieving self-labeling. Nevertheless, randomly augmenting images and especially random cropping can result in an image that is semantically very distant from the original and therefore leads to false labeling, hence undermining the efficacy of the methods. In this research, two novel parameterized cropping methods are introduced that increase the robustness of self-labeling and consequently increase the efficacy. The results show that the use of these methods significantly improves the accuracy of the model by between 2.7\% and 12.4\% on the downstream task of classifying CIFAR-10, depending on the crop size compared to that of the non-parameterized random cropping method.\footnote{The code for this research is available at \url{https://github.com/mhassan1a/Gaussian-Centered-Cropping}}
\end{abstract}    

\section{Introduction}
\label{sec:intro}

Deep learning models are data-hungry. Fortunately, in computer vision, there is plenty of data available. However, most of this data lacks task-specific labeling, which renders it out of reach for traditional supervised learning methods. Moreover, annotating large datasets is a task that is labor-intensive and, thus, almost unattainable in the case of datasets containing billions of data points. To give an example, it took 49,000 workers from Amazon Mechanical Turk from 167 countries approximately three years to annotate ImageNet, which consists of around 15 million images (Fei-Fei Li \cite{feilee}). Self-supervised learning methods (SSL) have proven to be a very effective solution to this problem, as they can learn useful representation directly from unlabeled data. Among the most promising SSL approaches is contrastive learning (CL), which takes advantage of some of the fundamental principles of knowledge representation (Haykin and Simon \cite{haykin2009neural}); These principles dictate that: firstly, similar inputs should typically yield close representations and secondly, inputs from different classes should produce distant representations. \\

Building upon these principles, CL methods use objective functions such as noise contrast estimation (Gutmann \etal\cite{infoNCE_2012}) to train the model to produce similar representations for instances of the same class and distinct representations for instances of different classes. CL methods achieve self-labeling by relying on augmentation techniques to produce random crops of the same image and treat them as instances of the same class (i.e. positive pairs). However, random cropping can result in transformations that do not preserve the true label (Shorten \etal\cite{shorten2019survey}), which leads to ``false labeling" also known as false positive, as shown in Fig. \ref{fig:fig10}. This problem poses a significant challenge as it may significantly hinder the training process and undermine the performance of the model.\\
    
Given the importance of cropping and the challenges posed by relying on random crops, it is clear that a shift from a purely stochastic cropping process to a more controllable one is necessary. Hence, the goal should be to propose a new approach that has the following properties:

\begin{itemize}
\addtolength{\itemindent}{0.2cm}
    \item \textbf{Consistency:} Eliminating false positives, thus improving the reliability of the crops as self-labels.
    \item \textbf{Relevance:}: Ensuring that crops are more likely to contain important features of objects, and so enriching the semantic content and informativeness of the generated views.
    \item \textbf{Flexibility:} Enabling the fine-tuning of the cropping process, ensuring optimal performance in various scenarios. 
\end{itemize}

To address these needs, this research introduces novel methods that sample crops from a parameterized Gaussian distribution with adjustable parameters. These methods aim to improve the effectiveness of contrastive learning methods by optimizing the cropping process to achieve reliable self-labeling. The contributions of this research are as follows: 
\begin{enumerate}
\addtolength{\itemindent}{1cm}
    \item The introduction of Gaussian-Centered Cropping (GCC) which is a parametric cropping method that samples crops from a Gaussian distribution with a fixed mean at the center of the image. This method is proven to outperform RandomCrop\footnote{https://pytorch.org/vision/stable/transforms.html} when used on centered datasets.
    \item The introduction of Multi-Object Gaussian-Centered Cropping (MGCC) which extending the GCC approach, MGCC takes into account the existence of multiple objects or regions of interest within the image or the case where the object is not centered. In this method, the mean of the distribution is not fixed, but is rather sampled from a uniform distribution around the center of the image. As a result, MGCC is capable of preventing the sampling of crops that are not semantically related without requiring semantic information, thus avoiding the need for resource-intensive computations or deep learning-based methods in the cropping process.
\end{enumerate}

\section{Related Work}
\label{sec:related_work}

%\subsection*{\textbf{Random Crop}}
In the framework of supervised learning, random crop as an image augmentation method is often considered a non-label-preserving transformation (Shorten \etal\cite{shorten2019survey}). This is because random cropping can significantly alter the content of the image, potentially deforming the object or scene depicted and leading to a loss of semantic information related to the pre-assigned label. However, in the context of self-supervised learning, random cropping has been proven crucial for the generation of different views of the same image, which are used as a form of self-labeling. Consequently, substantial efforts have been made in recent years to develop cropping methods that can produce semantically related views (i.e., positive pairs) that are sufficiently strong enough that the model is able to learn useful representation from them, while mitigating the risk of introducing unrelated views as positive pairs (i.e., false positives).\\

The most signification work in this regard is \textbf{Multi-Crop Strategy} (Caron \etal\cite{swav_2020}), which has been widely adopted by researchers (Caron \etal\cite{swav_2020}; Chen \etal\cite{fast_moco_2022}; Caron \etal\cite{dino_2021}; Pang \etal \cite{pang2022unsupervised}; Yeh \etal\cite{yeh2022decoupled}). This strategy involves creating multiple views of the image, including a number of global views (covering more than 50\% of the image) and some local views (covering less than 50\%). The use of smaller views increases the total number of views while maintaining the same computational cost (Caron \etal\cite{swav_2020}). Comparing smaller views to larger ones has been found to significantly boost the performance (Caron \etal\cite{swav_2020}), which is considered one of the strategy's greatest strengths. Nevertheless, Multi-Crop Strategy generates views using a non-parameterized random cropping method, which makes it prone to introducing false positives.\\

This challenge is addressed by \textbf{Object-Aware Cropping} (Mishra \etal\cite{object_aware_2021}) where a self-supervised object proposal network suggests a number of potential regions of interest (ROIs). Then an ROI is selected at random for the first view. The second view is obtained by enlarging the box that envelopes the first ROI and randomly cropping it. Nevertheless, one potential concern regarding this approach is the overhead computation required for the production of the first view, which is also very likely to be affected by the performance of the object proposal network. While another issue is that while the approach would reduce the likelihood of false positives, it still relies on non-parameterized random cropping methods to obtain the second view.\\

\textbf{Semantic-aware Localization} (Peng \etal\cite{Peng_2022}) is proposed as a solution to the object localization problem that does not require an object proposal network. In this method, a heat map, which is created from the features learned by the model, is used to infer the location of the objects in the image. After this, the views are created using RandomCrop at these locations. However, similar to Object-Aware Cropping (Mishra \etal\cite{object_aware_2021}) there is an overhead computational cost; the model needs to be trained using RandomCrop for a number of epochs to obtain the features that are required for the creation of the heat maps. The use of RandomCrop on top of the bounding box is believed to produce easy views for the loss function (Peng \etal\cite{Peng_2022}). The authors also propose \textbf{Contrastive Crop}: a method in which the views are samples from the bounding box, proposed by the Semantic-Aware Localization method, using a parameterized Beta distribution, the parameters of which are tuned as hyper-parameters (Peng \etal\cite{Peng_2022}).\\

Based on the work of Peng \etal\cite{Peng_2022}, this research and development project introduces a new method that does not require additional computational overhead costs. It can be used as a standalone views generator, or it can be integrated into Multi-Crop Strategy as a replacement for non-parametric random crop methods.
\section{Approach}
\label{sec:Approach}
In this section, the proposed parameterized cropping methods are presented in detail, as well as a comprehensive description of the different elements of this study.

\subsection{Methods}

The first method is \textbf{Gaussian-Centered Cropping (GCC)}, which creates two views around the center of the image. The centers of these views are sampled from a two-dimensional normal distribution, the mean of which is fixed at the center of the image. The variance of the distribution is defined as the product of a scaling factor, which is denoted as \textbf{$\alpha$}, and the dimensions of the image as width and height. By adjusting the value of \textbf{$\alpha$}, the variance of the distribution is controlled, hence the distance between the sampled views.  Algorithm \ref{algo:gauss_centered_cropping} provides a detailed description of the procedure.
Another variant of GCC is \textbf{Corrected Gaussian-Centered Cropping (CGCC)}, which prevents crops outside the image boundaries by moving the sampled center of the crop towards the center of the image until the entire view is fully contained within the boundary of the image.

\begin{algorithm}
\caption{Gaussian-Centered Cropping (GCC)}
\label{algo:gauss_centered_cropping}
\begin{algorithmic}[1]
\Function{GCC}{$\text{image}, \alpha, \text{crop\_size}$}
    \Statex \textbf{Input:} $\text{image}$, $\alpha$, $\text{crop\_size}$,
    \Statex \textbf{Output:} $\text{views}$
    
    \State $w, h \gets \text{image.shape}$
    \State $w_c, h_c \gets \sqrt{\text{crop\_size}} \times (w, h)$
    
    \State $\bm{\mu} \gets \left[ \frac{w}{2}, \frac{h}{2} \right]$
    \State $\bm{\Sigma} \gets \text{np.diag}\left( [\alpha w, \alpha h] \right)$
    \State $\text{centers} \gets \text{random.multivariate\_normal}(\bm{\mu}, \bm{\Sigma})$
    \State $\text{views} \gets \left[\text{ }\right]$
    
    \For{center in centers}
        \State $x, y \gets \text{center}$
        \State $l \gets x - \frac{w_c}{2}$
        \State $r \gets l + w_c$
        \State $t \gets y - \frac{h_c}{2}$
        \State $b \gets t + h_c$
        \State $view \gets \text{image}[t:b, l:r]$
        \State $\text{views.append(view)}$
    \EndFor
    
    \State \textbf{return} $\text{views}$
\EndFunction
\end{algorithmic}
\end{algorithm}

The third method is \textbf{Multi-Object Gaussian-Centered Cropping (MGCC)}. The only difference between MGCC, as described in Algorithm \ref{algo:multi_object_Gaussian_centered_cropping}, and GCC lies in how the centers of the crops are sampled. In MGCC, the centers of the crops are sampled from a multivariate normal distribution. Unlike GCC, the mean of the distribution is not fixed to the center of the image, but it is sampled instead from a uniform distribution around the center of the image, the boundaries of this distribution defined as hyperparameters a and b. This approach is expected to enhance self-supervised learning from image datasets in which either objects are not centered or multiple objects are present in the same image or both. Another variant of MGCC is \textbf{Multi-Object Corrected Gaussian-Centered Cropping (MCGCC)}, which is analogous to CGCC. 

\begin{algorithm}
\caption{Multi-Object Gaussian-Centered Cropping (MGCC)}
\label{algo:multi_object_Gaussian_centered_cropping}
\begin{algorithmic}[1]
\Function{MGCC}{$\text{image}, \alpha, \text{crop\_size}, (a, b)$}
    \Statex \textbf{Input:} $\text{image}$, $\alpha$, $\text{crop\_size}$, $\text{(a, b)}$
    \Statex \textbf{Output:} $\text{views}$
    \State $w, h \gets \text{image.shape}$
    \State $w_c, h_c \gets \sqrt{\text{crop\_size}} \times (w, h)$
    \State $\mu_x \gets \text{random.uniform}(a \times w, b \times w)$
    \State $\mu_y \gets \text{random.uniform}(a \times h, b \times h)$
    \State $\bm{\mu} \gets [\mu_x, \mu_y]$
    \State $\bm{\Sigma} \gets \text{np.diag}\left( [\alpha w, \alpha h] \right)$
    \State $\text{centers} \gets \text{random.multivariate\_normal}(\bm{\mu}, \bm{\Sigma})$
    \State $\text{views} \gets \left[\text{ }\right]$
    
    \For{center in centers}
        \State $x, y \gets \text{center}$
        \State $l \gets x - \frac{w_c}{2}$
        \State $r \gets l + w_c$
        \State $t \gets y - \frac{h_c}{2}$
        \State $b \gets t + h_c$
        \State $view \gets \text{image}[t:b, l:r]$
        \State $\text{views.append(view)}$
    \EndFor
    
    \State \textbf{return} $\text{views}$
\EndFunction
\end{algorithmic}
\end{algorithm}

\subsection{Experiment Setup}
The experiment is divided into two main stages:

\begin{itemize}
    \item \textbf{Pretraining on a Pretext Task}:
During this stage, the model undergoes pretraining on an embeddings prediction task, whereby it learns features by trying to encode classes in an arbitrary lower dimensional space.

    \item \textbf{Evaluation on Transfer Learning Task}:
After pretraining the model is evaluated on a transfer learning task, in which a new classifier is trained to predict class labels to assess the quality of the features learned during the pretraining stage.
\end{itemize}

\subsection{Datasets}
In this project, the following datasets are utilized for the pretraining and evaluation of the models:
\begin{itemize}
    \item \textbf{CIFAR-10} \cite{cifar10} is a benchmark dataset for image classification tasks. It consists of 60,000 32x32 pixels colored images across 10 classes, with 6000 images per class.

    \item \textbf{TinyImageNet} \cite{tinyimagenet} is a subset of the ImageNet dataset, containing 200 classes with 500 training images and 50 validation images per class. Each image is 64x64 in size.
    
    \item\textbf{ImageNet64} \cite{imagenet64} is a variant of the ImageNet dataset where each image is resized to 64x64, containing the same classes as the original ImageNet dataset but with reduced image resolution.
\end{itemize}

\subsection{Models}

For pretraining on CIFAR-10 and TinyImageNet, a modified ResNet-18 was employed as the backbone model. Whereas when the pretraining was carried out on ImageNet64, ResNet-34 was utilized. The selection of these models was based on related research by (Peng \etal\cite{Peng_2022}) as well as the relative sizes of the datasets. The smaller ResNet-18 is well suited for the relatively smaller datasets of CIFAR-10 and TinyImageNet, providing sufficient capacity for learning without excessive computational overhead. On the other hand, the larger ResNet-34 is more appropriate for the more extensive and diverse ImageNet64 dataset, offering enhanced representational power to handle the increased complexity and variability. The modifications to both networks were the same, and are as follows.
\begin{itemize}
    \item The size of the kernel of the first convolution layer was reduced to (3x3) since the images are small.
    \item The max-pooling layer was removed for the same reason.
    \item The output of the classifier was increased to an arbitrary embedding dimension which can be considered a hyperparameter.
\end{itemize}

After pretraining, the classifier was removed and the weights of the network were frozen. A new linear classifier was added for linear evaluation, with the classifier consisting of a single linear layer.

\subsection{Training and Loss Function}
\label{NT-Xent}
The training framework follows a methodology similar to that of SimCLR \cite{simCLR2020}. The model is trained to learn class representations in the embedding dimension from the pretext task of identifying similar images. This is achieved using Normalized Temperature-Scaled Cross-Entropy (NT-Xent) loss\cite{nxtloss}, a variant of the InfoNCE\cite{infoNCE_2012} loss commonly employed in contrastive learning frameworks such as SimCLR\cite{simCLR2020} .\\

NT-Xent loss plays a pivotal role in encouraging the model to bring together representations of positive pairs, which consist of augmented versions of the same image, while simultaneously pushing apart representations of negative pairs, which correspond to representations of different images. Mathematically, the NT-Xent loss is defined as in the following equation \eqref{ntxet}:

\begin{equation}\label{ntxet}
\mathcal{L}(z_i, z_j) = -\log \frac{\exp(\frac{\text{sim}(z_i, z_j)}{ \tau})}{\sum_{k=1}^{2N} \mathbf{1}_{[k \neq i]} \exp(\frac{\text{sim}(z_i, z_k)}{\tau})}
\end{equation}

whereby:
\begin{itemize}
    \item $z_i$ and $z_j$ are the representations of augmented views of the same image (positive pair),
    \item $\text{sim}(z_i, z_j)$ denotes the cosine similarity between $z_i$ and $z_j$,
    \item $\tau$ is the temperature parameter that scales the logits, and
    \item $\mathbf{1}_{[k \neq i]}$ is the indicator function that is 1 if $k \neq i$ and 0 otherwise.
    \item ${2N}$ denotes twice the batch size $N$, where $N$ represents the batch size, considering that two views are generated for each image.
\end{itemize}

\subsection{Augmentations and Preprocessing}
In addition to the cropping method, the following augmentations and preprocessing were applied to the images:

\begin{itemize}
    \item Random horizontal flip
    \item Gaussian blur with a kernel size of (3, 3)
    \item Normalization by standardization
\end{itemize}

\subsection{Evaluation Criteria}
\label{subsection:evaluation}
Contrastive learning methods are typically assessed based on their ability to learn meaningful representations that can be effectively transferred to downstream tasks. The primary evaluation metrics include:

\begin{itemize}
    \item \textbf{Linear Evaluation Protocol (LEP):} This involves training a linear classifier on the representations learned by the contrastive learning model (He \etal\cite{moco_2020}; Chen et al. \cite{simCLR2020}). The performance of this classifier on a validation or test set is indicative of the quality of the learned representations.
    \item \textbf{Transfer Learning:} The learned representations are transferred to other tasks or datasets to evaluate their generalizability and robustness (He \etal\cite{moco_2020}). The performance on these new tasks provides insight into the versatility of the model.
    \item \textbf{Clustering Accuracy:} This metric assesses how well the learned representations can cluster similar data points (Caron \etal\cite{deep_cluster_2018, swav_2020}). The high clustering accuracy indicates that the model has captured meaningful similarities and differences in the data.
\end{itemize}

In this project, a \textbf{Linear Evaluation Protocol (LEP)} was followed, and the evaluation can be described in \ref{subsection:evalprot}. The selection of LEP is driven by its straightforwardness and prevalence in the field. 

\subsection{Evaluation Protocol}
\label{subsection:evalprot}
The evaluation protocol consists of several steps which can be summarized as follows:
\begin{itemize}
    \item The model undergoes pretraining in an unsupervised setting for a predefined number of epochs.
    \item Following pretraining, the weights of the backbone are frozen and the linear layer is removed.
    \item Subsequently, a linear classifier is trained on top of the frozen feature maps in a supervised learning setting.
    \item Finally, the performance of this classifier on the test set is used to compare different pretraining settings.
    \item In addition, to ensure that the performance genuinely reflects the quality of pretraining, rather than the complexity of the linear classifier, the model was evaluated without pretraining to establish a baseline.
\end{itemize}
\section{RESULTS}
\label{sec:evaluation}
\subsection{\textbf{Results on CIFAR-10}}
To investigate the impact of changing the variance on the performance of the model when pretrained on an acurated dataset of centered images, the model underwent an initial unsupervised pretraining phase on the CIFAR-10 dataset. This was followed by an evaluation according to the protocol described in Sec. \ref{subsection:evaluation}. The whole experiment was repeated four times for each value of  $\alpha$ (i.e., variance control parameters) and the accuracy of the downstream classifier was recorded along with the mean and standard deviation. This assessment was carried out after 200 epochs and was repeated for different crop sizes, allowing the examination of the potential synergies between the value of $\alpha$, the cropping method and the crop size. Subsequently, the model was pretrained for 500 epochs using selected values for $\alpha$, allowing a comparative study of the different methods with various crop sizes. \\

\begin{figure}[ht]
    \begin{subfigure}{\linewidth}
        \centering
        \includegraphics[width=1\linewidth]{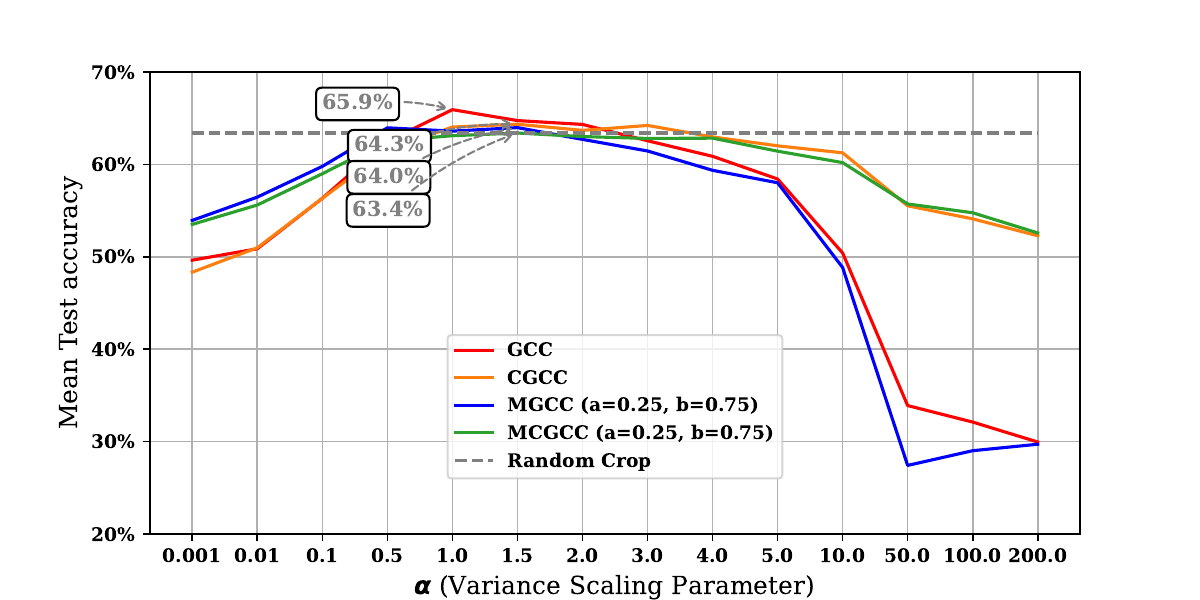}
        \caption{Crop size of 20\%.}
        \label{fig:5a}
    \end{subfigure}

    \begin{subfigure}{\linewidth}
        \centering
        \includegraphics[width=1\linewidth]{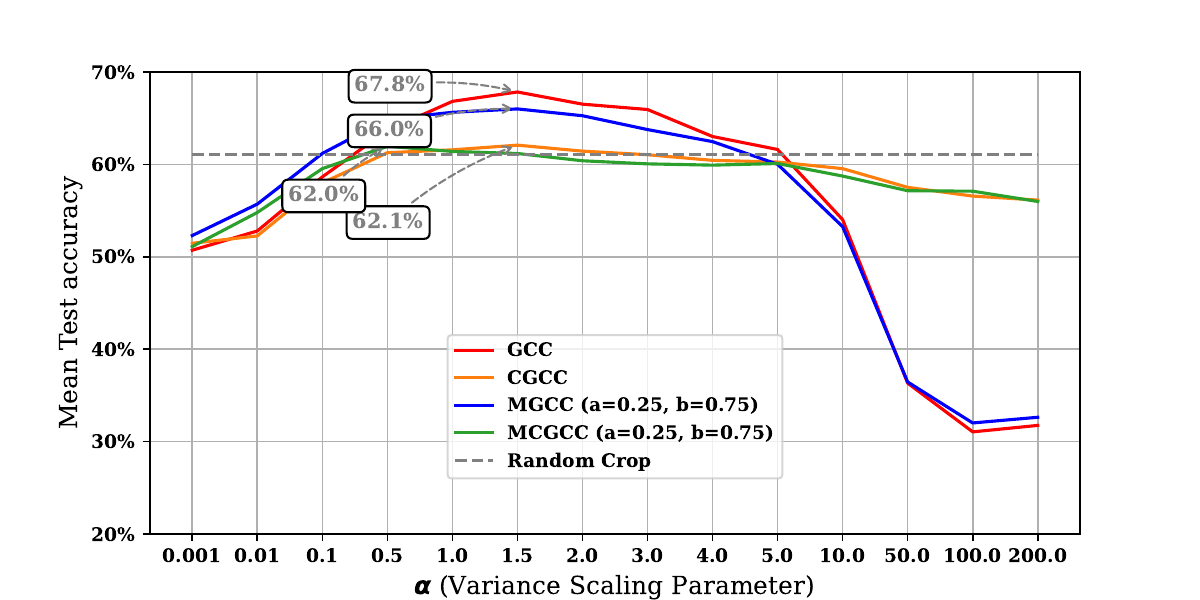}
        \caption{Crop size of 40\%.}
        \label{fig:5b}
    \end{subfigure}

    \begin{subfigure}{\linewidth}
        \centering
        \includegraphics[width=1\linewidth]{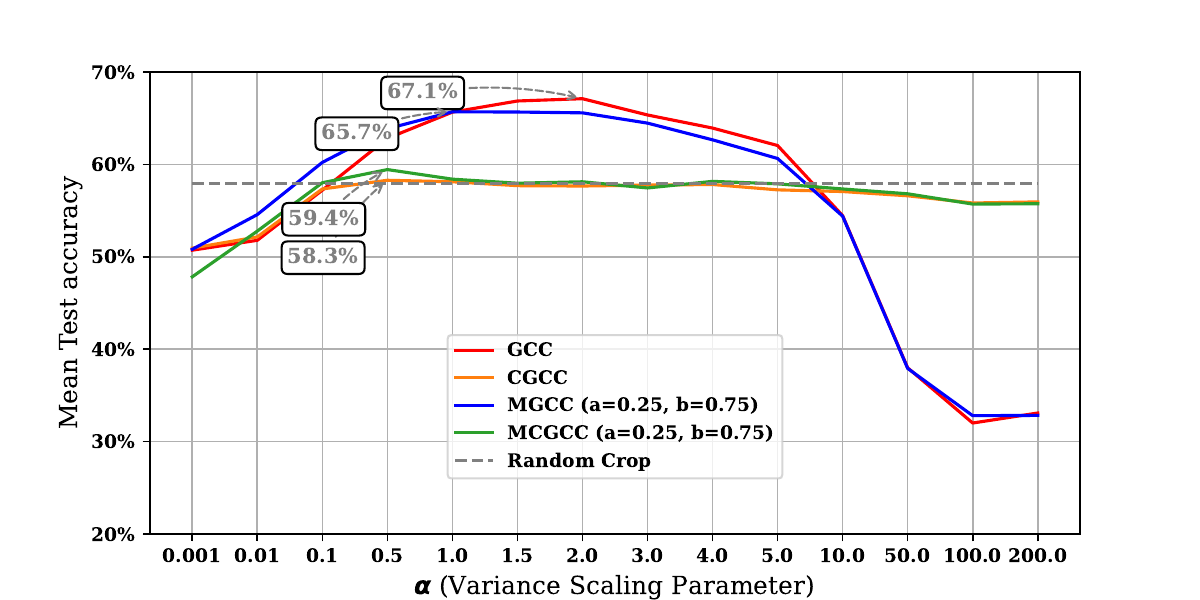}
        \caption{Crop size of 60\%.}
        \label{fig:5c}
    \end{subfigure}

    \begin{subfigure}{\linewidth}
        \centering
        \includegraphics[width=1\linewidth]{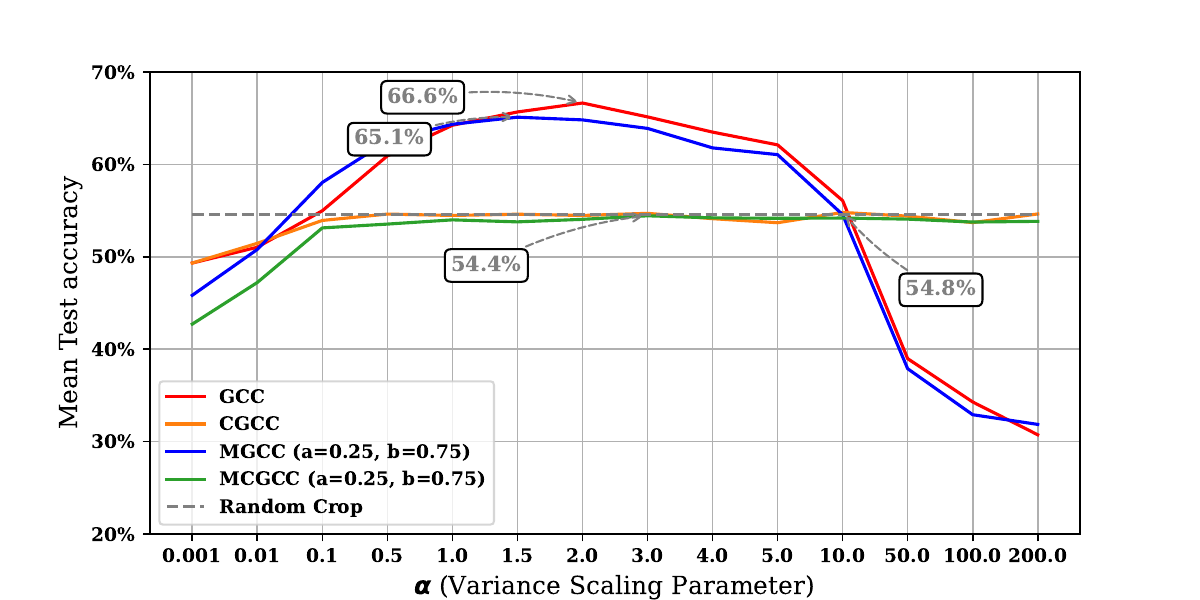}
        \caption{Crop size of 80\%.}
        \label{fig:5d}
    \end{subfigure}
    \caption{Effect of the variance scaling parameter alpha on the performance of each of the proposed methods. Subplots show different crop sizes. The pretraining was performed on CIFAR-10 for 200 epochs.}
    \label{fig:fig1}
\end{figure}

Pretraining with \textbf{GCC} and \textbf{MGCC} yields comparable results. Fig. \ref{fig:fig1} reveals a strong relationship between $\alpha$ (i.e., variance) and the performance of the model. On the left-hand side of each graph, there is a strong positive correlation between variance and accuracy. As the variance increases, so does the performance of the model until it reaches a peak point. However, beyond this peak point, any further increase in the value of the variance is associated with a deterioration in performance. This pattern suggests an optimal value for the variance that maximizes the performance of the downstream model, due to the better quality of the features learned in the pretraining stage.\\ 

\begin{figure}[ht]
            \centering
            \includegraphics[width=1\linewidth]{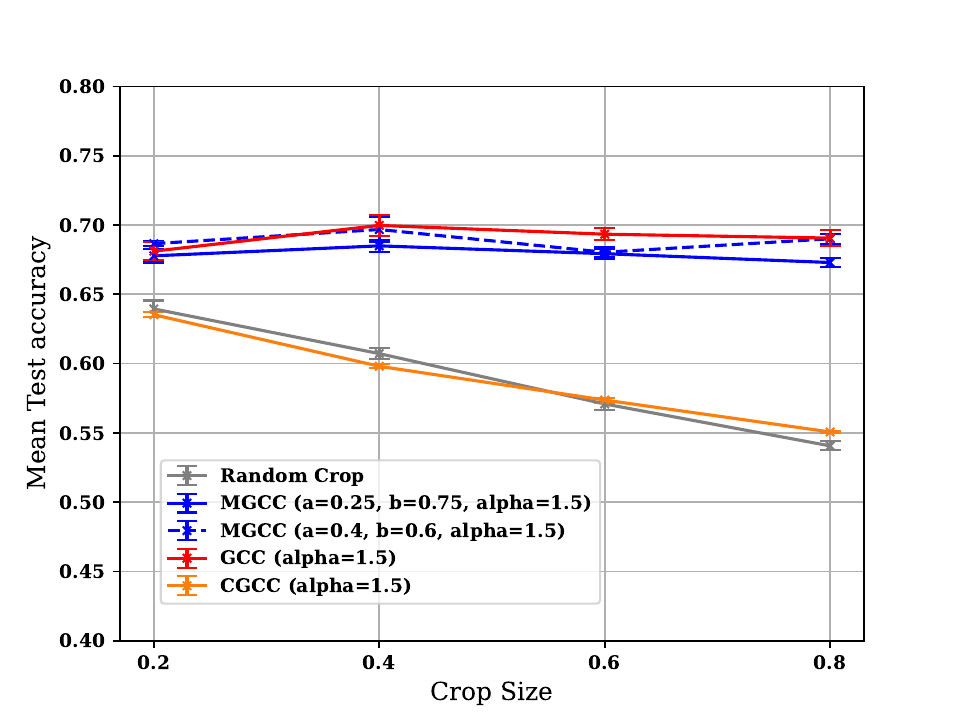}
            \caption{Methods performance regarding crop size. Showing the accuracy of the downstream classifier after pretraining on CIFAR-10 for 500 epochs with varying crop size.}
            \label{fig:fig2}
        \end{figure}

Furthermore, in Fig. \ref{fig:fig1} the behavior of the two corrected methods namely \textbf{CGCC} and \textbf{MCGCC} can be compared with that of \textbf{GCC} and \textbf{MGCC} where the performance of the corrected method plateaus and does not exceed that of RandomCrop. They are also more stable with high variance. Table. \ref{tab:crop_comparison} draws a comparison between the performance of the different methods and highlights the advantages of \textbf{GCC} and \textbf{MGCC} over all other methods. Finally, to examine the effect of the pretraining time on the results, the experiments were repeated with a higher number of epochs equal to 500. Fig. \ref{fig:fig2} shows that while more pretraining yields slightly better performance, the overall patterns remain the same. 

\begin{table}[htbp]
    \centering
    \caption{Pretraining on CIFAR10 for 200 epochs. Peak accuracy comparison between GCC, CGCC, MGCC, MCGCC, and RandomCrop across different crop sizes (values are in percentages).}
    \label{tab:crop_comparison}
    \begin{tabular}{@{}cccccc@{}}
        \toprule
        & \multicolumn{4}{c}{\textbf{Crop Size}} \\
        \cmidrule(lr){2-5}
        \textbf{Method} & \textbf{0.2} & \textbf{0.4} & \textbf{0.6} & \textbf{0.8} \\
        \midrule
        %\rowcolor[HTML]{EFEFEF}
        \textbf{GCC} & \textbf{65.9} & \textbf{67.8} & \textbf{67.1} & \textbf{66.6} \\
        \textbf{MGCC} & 64.0 & 66.0 & 65.7 & 65.1 \\
        %\rowcolor[HTML]{EFEFEF}
        \textbf{CGCC} & 64.3 & 62.1 & 58.3 & 54.8 \\
        \textbf{MCGCC} & 63.4 & 62.0 & 59.4 & 54.4 \\
        %\rowcolor[HTML]{EFEFEF}
        \textbf{RandomCrop} & 63.2 & 61.1 & 57.6 & 54.2 \\
        \bottomrule
    \end{tabular}
\end{table}

\subsection{\textbf{Results on TinyImageNet}}
To assess the generalizability of the results beyond CIFAR-10, a training routine was conducted using GCC and MGCC methods with selected hyperparameters on TinyImageNet \cite{tinyimagenet}. The performance was compared with that of RandomCrop. The decision to exclude the corrected methods was based on their performance, which did not exceed that of RandomCrop in previous experiments. Furthermore, the selection of hyperparameters was guided by insights from earlier experiments.\\

\begin{figure}[ht]
            \centering
            \includegraphics[width=1\linewidth]{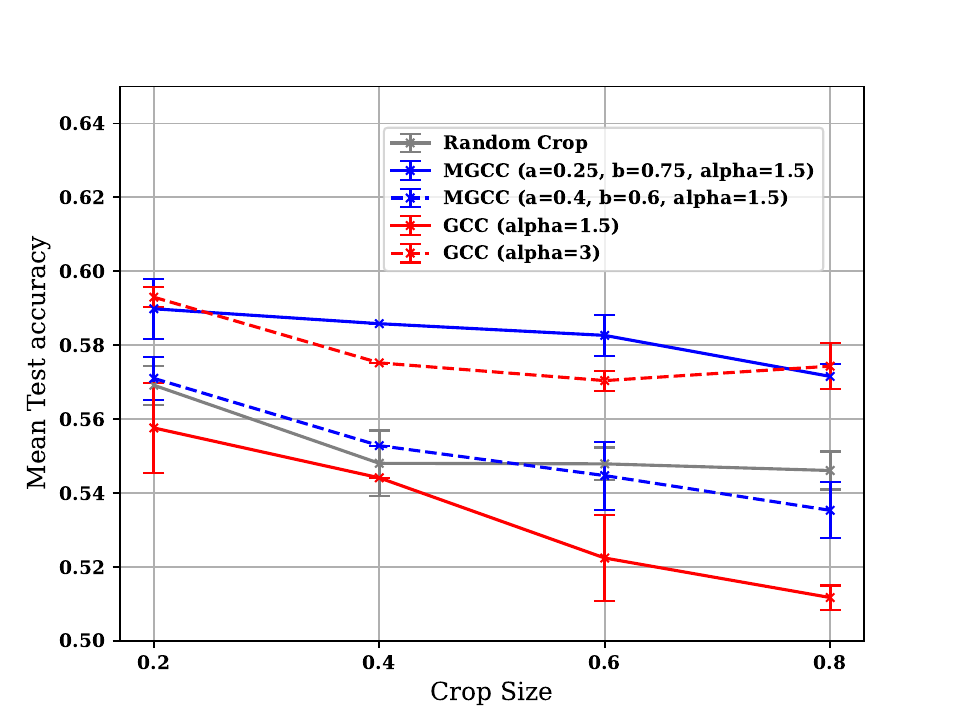}
            \caption{Methods performance regarding crop size. Showing the accuracy of the downstream classifier after pretraining on TinyImageNet for 200 epochs with varying crop size.}
            \label{fig:fig7}
        \end{figure}

As illustrated in Fig. \ref{fig:fig7}, both GCC and MGCC outperformed RandomCrop when the correct parameters were chosen. In particular, MGCC outperformed GCC in crop sizes of 40\% and 60\%. This result aligns with expectations given that, unlike CIFAR-10 images, which typically contain a single centered object, TinyImageNet images can contain multiple objects. Thus, the multi-object capability of MGCC proves advantageous in this context.

\subsection{\textbf{Results on ImageNet64}}
\begin{figure}[ht]
            \centering
            \includegraphics[width=1\linewidth]{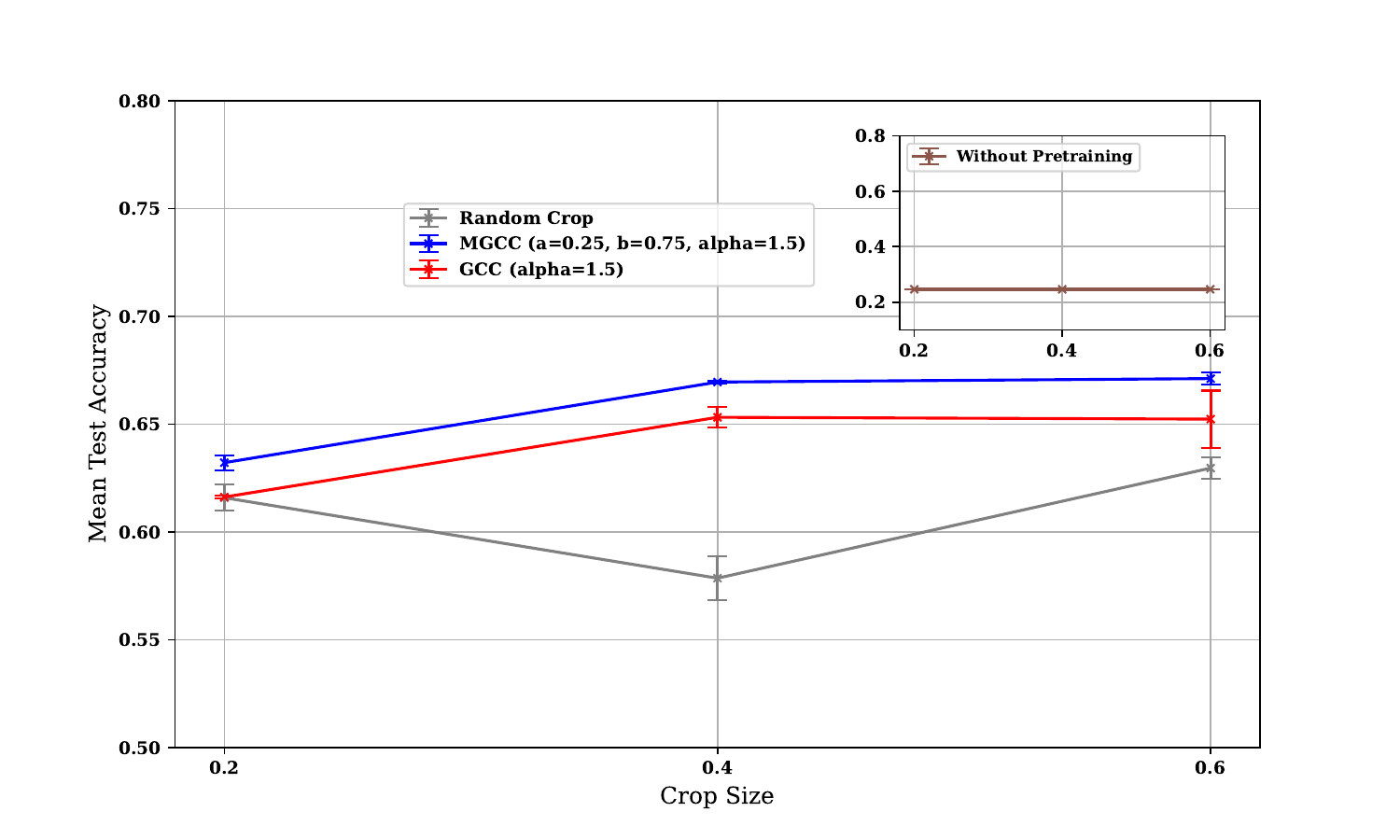}
            \caption{Methods performance regarding crop size. Showing the accuracy of the downstream classifier after pretraining on ImageNet64 for 200 epochs with varying crop size.}
            \label{fig:fig8}
        \end{figure}

A final evaluation was carried out using the ImageNet64 dataset \cite{imagenet64}. The evaluation follows a methodology similar to that of the TinyImageNet assessment. The results, presented in Fig. \ref{fig:fig8}, demonstrate the superiority of GCC and MGCC over RandomCrop. In particular, MGCC exhibits particularly promising results, consistently outperforming GCC and showing enhanced performance across various crop sizes. These findings underscore the efficacy of both GCC and MGCC in improving model performance, especially when faced with larger and more complex datasets like ImageNet64. The significant improvement observed with MGCC is consistent with that of TinyImageNet and highlights the potential of MGCC as a robust augmentation technique for complex image recognition tasks.

\section{Discussion}

On CIFAR-10, it was observed that the performance gap between GCC and MGCC and the other methods widens as the crop size increases. The smallest performance difference occurs with a crop size of 20\%. This could be due to the fact that smaller crops introduce more invariance and are more likely to produce weaker false positives than larger crops, as objects often share small-scale features. Robust classification relies not only on capturing local features, but also on understanding the relationships between these features within a global context. This hypothesis is further supported by the fact that the GCC and MGCC methods exhibited peak performance at a crop size of 40\% , followed by a slight decrease in performance at larger crop sizes.\\

This suggests that a trade-off is necessary. Smaller crop sizes lack sufficient global context, leading to weaker high-level features. In contrast, larger crop sizes reduce the invariance introduced by cropping and hence result in stronger false positives with a higher likelihood to include divergent features. Therefore, careful selection of crop size is crucial to achieve an optimal balance between these conflicting requirements. The results also show that the optimal crop sizes differed from one dataset to another, highlighting that the ideal crop size depends on the characteristics of the dataset, particularly the relative size of objects in relation to the dimensions of the image.\\

Interestingly, GCC and MGCC demonstrated superior performance compared to their corrected variants, CGCC and MCGCC, which might seem counterintuitive since these methods can produce crops that do not fully fit within the boundaries of the image, a behavior that can result in the loss of some semantic information. Conceptually, one possible explanation is that these methods might function similarly to the Multi-Crop Strategy, particularly when the crop center is far from the image center. In the Multi-Crop Strategy, both local (small-scale) and global (large-scale) features are encoded in a manner that promotes consistent representations across different scales.\\

\begin{figure}[ht]
    \centering
    \begin{subfigure}{\linewidth}
        \centering
        \includegraphics[width=0.7\linewidth, height=0.15\textheight]{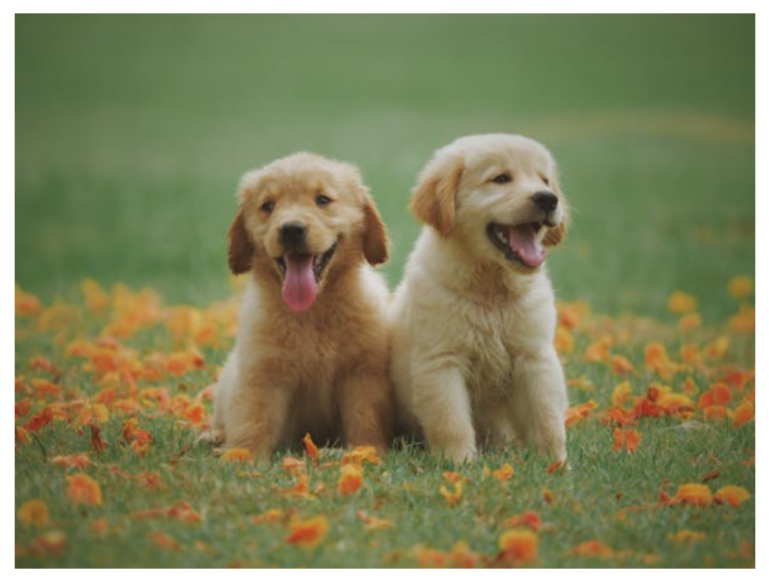}
        \centering\caption{\centering Original image. Retrieved from: \href{http://carnimed.de/blog/ein-welpe-zieht-ein-was-solltest-du-bei-der-fuetterung-von-welpen-beachten}{carnimed}\newline Copyright information not available.} 
        \label{fig:4a}
    \end{subfigure}

    \begin{subfigure}{\linewidth}
        \centering
        \begin{tikzpicture}
            \node[anchor=south west,inner sep=0] at (-3,-0.9) {\includegraphics[width=0.9\linewidth]{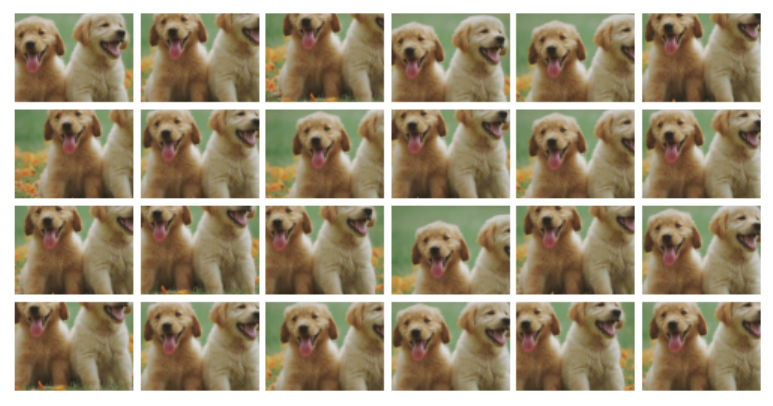}};
            \draw[red,thick,rounded corners] (2,2) rectangle (4.5,3);  % Adjust coordinates as needed
        \end{tikzpicture}
        \caption{GCC (alpha=1).}
        \label{fig:4c}
    \end{subfigure}

    \begin{subfigure}{\linewidth}
        \centering
        \begin{tikzpicture}
            \node[anchor=south west,inner sep=0] at (-3,-0.9) {\includegraphics[width=0.9\linewidth]{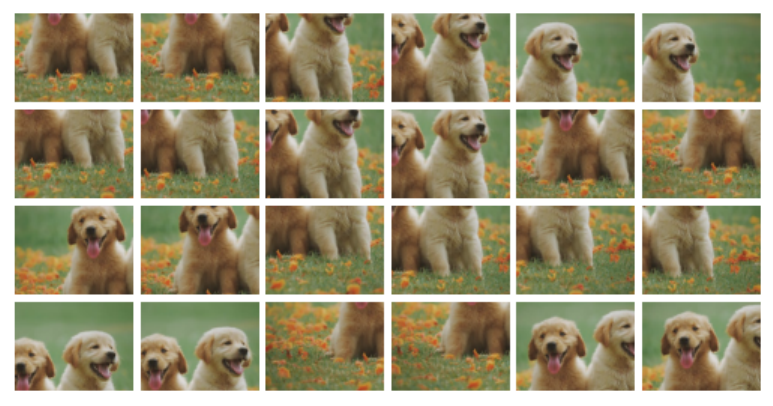}};
            \draw[red,thick,rounded corners] (2,2) rectangle (4.5,3);  % Adjust coordinates as needed
        \end{tikzpicture}
        \caption{MGCC (alpha=1, a=0.25, b=0.75)}
        \label{fig:4b}
    \end{subfigure}

    \begin{subfigure}{\linewidth}
        \centering
        \begin{tikzpicture}
            \node[anchor=south west,inner sep=0] at (-3,-0.9) {\includegraphics[width=0.9\linewidth]{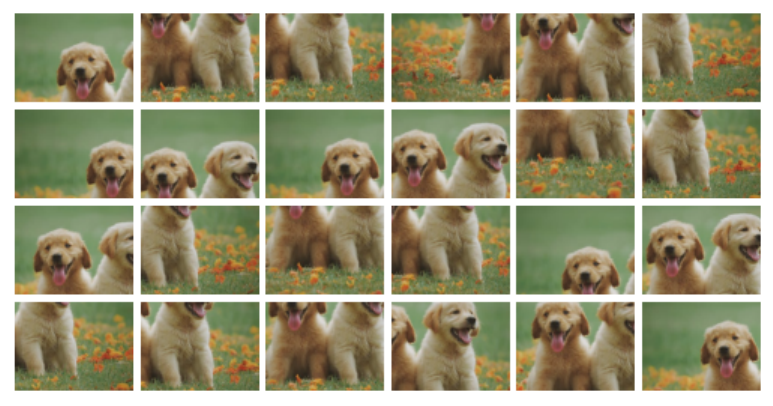}};
            \draw[red,thick,rounded corners] (2,2) rectangle (4.5,3);  % Adjust coordinates as needed
        \end{tikzpicture}
        \caption{RandomCrop}
        \label{fig:4d}
    \end{subfigure}

    \caption{The effect of the different cropping methods applied to a centered image with 20\% crop size. The crops are ordered from left to right each consecutive pair of images in a row represents a proposed positive pair, the red boxes indicate this pairwise relationship.}
    \label{fig:fig9}
\end{figure}
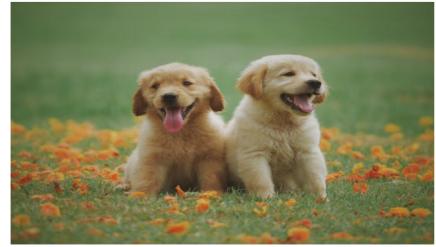
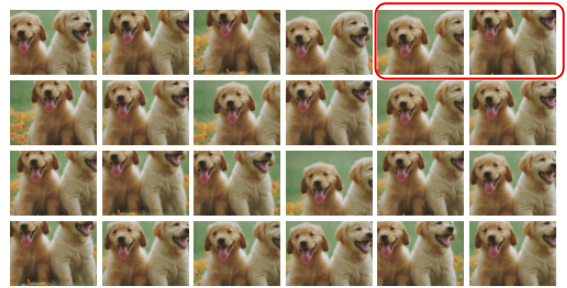
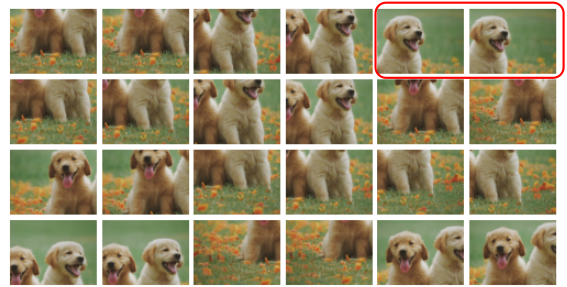
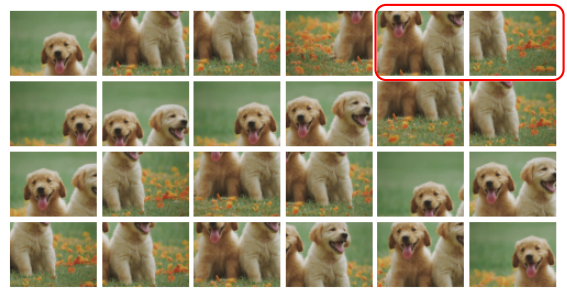

However, the aforementioned explanation is not supported by the results, as optimal performance was observed at low values for alpha. This suggests that the crop centers were close to the image center and required minimal padding, as shown in Fig. \ref{fig:fig9}. Consequently, the improved performance of GCC and MGCC may be attributed to their ability to effectively produce consistent, non-trivial positive pairs at optimal distances from each other, thus enabling the model to learn effectively without relying on extensive cropping or padding. It might also be attributed to a synergy between the nature of the underlying distribution, in this case the normal distribution, and the performance.\\

On TinyImageNet, both GCC and MGCC outperformed RandomCrop when appropriately parameterized, with MGCC showing superior performance in crop sizes 40\% and 60\%. These results align with expectations, as TinyImageNet images often contain multiple objects. The superior performance of MGCC can be attributed to its ability to handle multiple objects within a single image, thereby reducing false positives and increasing invariance. By generating crops that capture more diverse regions of the image, MGCC ensures that both local and global features are represented, leading to more robust learning.\\

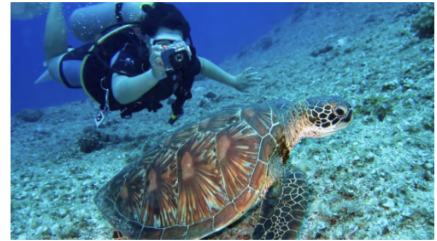
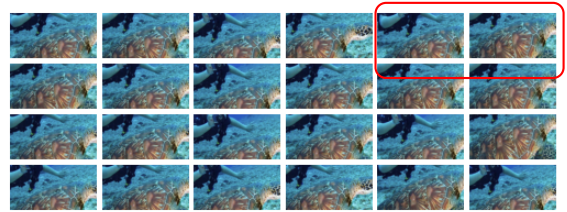
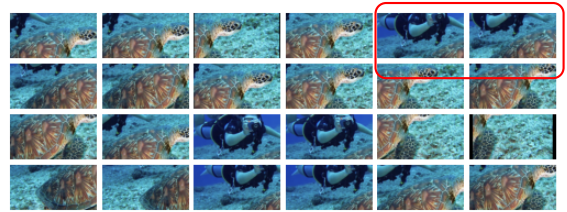
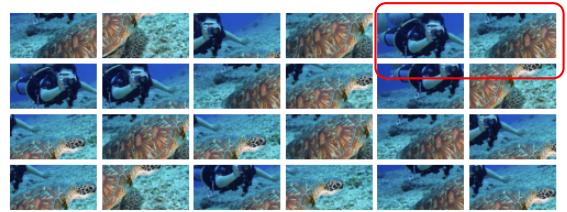
\begin{figure}[ht]
    \centering
    \begin{subfigure}{\linewidth}
        \centering
        \includegraphics[width=0.7\linewidth, height=0.15\textheight]{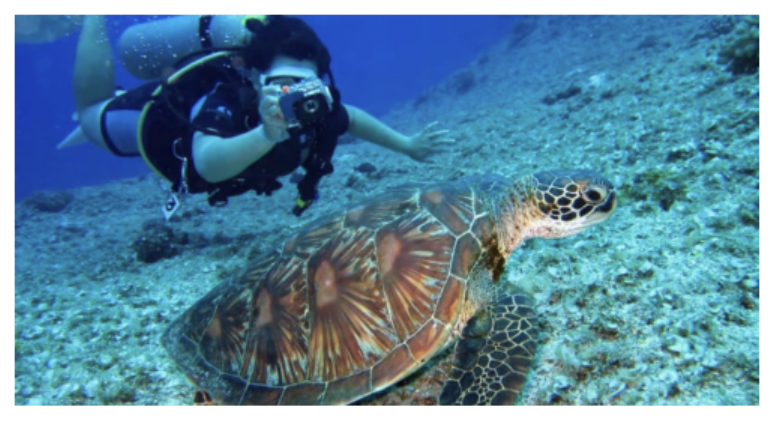}
        \caption{\centering Original image. Retrieved from: \href{https://www.reefsmagazine.com/reef-keeping/}{reefsmagazine}\newline Copyright information not available.}
        \label{fig:4a2}
    \end{subfigure}

    \begin{subfigure}{\linewidth}
        \centering
        \begin{tikzpicture}
            \node[anchor=south west,inner sep=0] at (-3,0.1) {\includegraphics[width=0.9\linewidth]{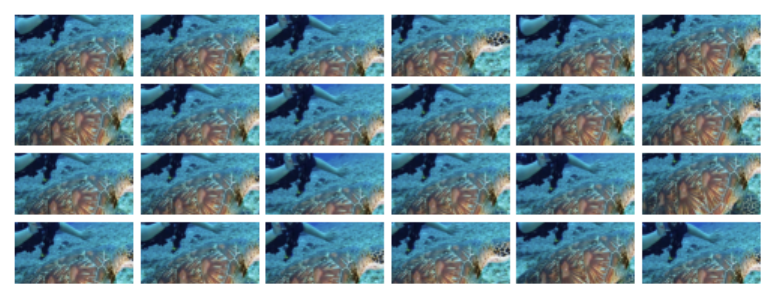}};
            \draw[red,thick,rounded corners] (2,2) rectangle (4.5,3);  % Adjust coordinates as needed
        \end{tikzpicture}
        \caption{GCC (alpha=1)}
        \label{fig:4c2}
    \end{subfigure}

    \begin{subfigure}{\linewidth}
        \centering
        \begin{tikzpicture}
            \node[anchor=south west,inner sep=0] at (-3,0.1) {\includegraphics[width=0.9\linewidth]{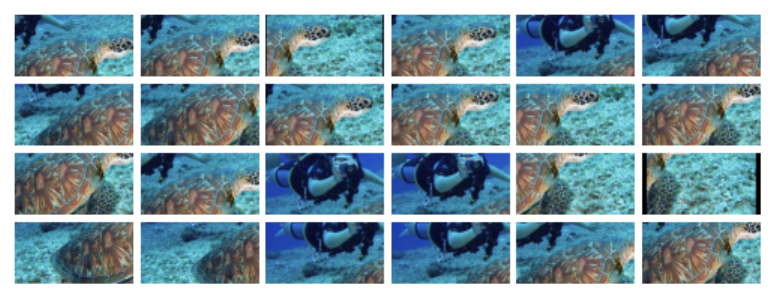}};
            \draw[red,thick,rounded corners] (2,2) rectangle (4.5,3);  % Adjust coordinates as needed
        \end{tikzpicture}
        \caption{MGCC (alpha=1, a=0.25, b=0.75)}
        \label{fig:4d2}
    \end{subfigure}

    \begin{subfigure}{\linewidth}
        \centering
        \begin{tikzpicture}
            \node[anchor=south west,inner sep=0] at (-3,0.1) {\includegraphics[width=0.9\linewidth]{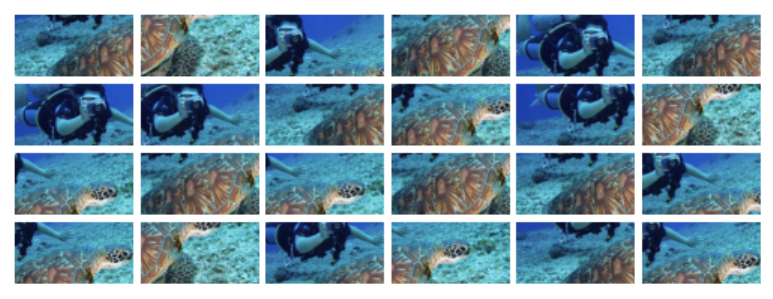}};
            \draw[red,thick,rounded corners] (2,2) rectangle (4.5,3);  % Adjust coordinates as needed
            
        \end{tikzpicture}
        \caption{RandomCrop}
        \label{fig:4d2}
    \end{subfigure}

    \caption{The effect of the different cropping methods applied to a non-centered image with multiple objects with 20\% crop size. The crops are ordered from left to right each consecutive pair of images in a row represents a proposed positive pair, the red boxes indicate this pairwise relationship.}
    \label{fig:fig10}
\end{figure}

Fig. \ref{fig:fig10}  illustrates how MGCC excels in generating crops that cover the entire image while maintaining an optimal distance between them. This is in contrast to the RandomCrop and GCC methods. RandomCrop tends to introduce false positives more frequently because it randomly selects crop regions without considering their relative distances. GCC, on the other hand, exhibits limitations based on the value of alpha. For small alpha values, which are required to achieve an optimal distance between crops as the results suggest, GCC crops are often concentrated around the center of the image. This concentration can result in inadequate coverage of the diverse features of the image. In contrast, for larger alpha values, the increased distance between crops can lead to more false positives and a deviation from the optimal cropping distances.\\

Finally, the computational efficiency of both MGCC and GCC is supported by their reliance on well-understood and optimized normal distribution sampling, along with simple padding when necessary. These straightforward requirements set them apart from approaches such as Object-Aware Cropping (Mishra et al. \cite{object_aware_2021}), which relies on an object proposal neural network, and Semantic-Aware Localization (Peng et al. \cite{Peng_2022}), which depends on the generation of heat maps. This efficient design not only reduces computational demands but also facilitates their integration into popular deep learning frameworks.

\section{Conclusion}
In conclusion, in this project, two novel cropping methods, GCC and MGCC, are proposed. Both methods have consistently outperformed their corrected counterparts (i.e., CGCC and MCGCC) and RandomCrop. However, the main disadvantage of GCC arises when dealing with arbitrary images in which the object is not centered or multiple objects are present. In such cases, GCC is prone to producing false positives. This limitation highlights the need for MGCC, which is designed to handle these situations more robustly. More research is needed, particularly with regard to finding optimal crop sizes. One potential direction is exploring layered cropping, whereby MGCC could be used to generate initial batches of crops, followed by GCC application on these batches to enhance performance.

{
    \small
    \bibliographystyle{ieeenat_fullname}
    \bibliography{main.bib}

\begin{thebibliography}{18}
\providecommand{\natexlab}[1]{#1}
\providecommand{\url}[1]{\texttt{#1}}
\expandafter\ifx\csname urlstyle\endcsname\relax
  \providecommand{\doi}[1]{doi: #1}\else
  \providecommand{\doi}{doi: \begingroup \urlstyle{rm}\Url}\fi

\bibitem[Caron et~al.(2018)Caron, Bojanowski, Joulin, and Douze]{deep_cluster_2018}
Mathilde Caron, Piotr Bojanowski, Armand Joulin, and Matthijs Douze.
\newblock \emph{Deep Clustering for Unsupervised Learning of Visual Features}, page 139–156.
\newblock Springer International Publishing, 2018.

\bibitem[Caron et~al.(2020)Caron, Misra, Mairal, Goyal, Bojanowski, and Joulin]{swav_2020}
Mathilde Caron, Ishan Misra, Julien Mairal, Priya Goyal, Piotr Bojanowski, and Armand Joulin.
\newblock Unsupervised learning of visual features by contrasting cluster assignments.
\newblock \emph{Advances in Neural Information Processing Systems}, 33:\penalty0 9912--9924, 2020.

\bibitem[Caron et~al.(2021)Caron, Touvron, Misra, J{\'e}gou, Mairal, Bojanowski, and Joulin]{dino_2021}
Mathilde Caron, Hugo Touvron, Ishan Misra, Herv{\'e} J{\'e}gou, Julien Mairal, Piotr Bojanowski, and Armand Joulin.
\newblock Emerging properties in self-supervised vision transformers.
\newblock In \emph{Proceedings of the IEEE/CVF International Conference on Computer Vision}, pages 9650--9660, 2021.

\bibitem[Chen et~al.(2020)Chen, Kornblith, Norouzi, and Hinton]{simCLR2020}
Ting Chen, Simon Kornblith, Mohammad Norouzi, and Geoffrey Hinton.
\newblock A simple framework for contrastive learning of visual representations.
\newblock In \emph{International Conference on Machine Learning}, pages 1597--1607. PMLR, 2020.

\bibitem[Ci et~al.(2022)Ci, Lin, Bai, and Ouyang]{fast_moco_2022}
Yuanzheng Ci, Chen Lin, Lei Bai, and Wanli Ouyang.
\newblock Fast-moco: Boost momentum-based contrastive learning with combinatorial patches.
\newblock In \emph{European Conference on Computer Vision}, pages 290--306. Springer, 2022.

\bibitem[Deng et~al.(2009)Deng, Dong, Socher, Li, Li, and Fei-Fei]{imagenet64}
Jia Deng, Wei Dong, Richard Socher, Li-Jia Li, Kai Li, and Li Fei-Fei.
\newblock Imagenet: A large-scale hierarchical image database.
\newblock In \emph{IEEE/CVF Conference on Computer Vision and Pattern Recognition (CVPR)}, 2009.

\bibitem[Fei-Fei and Deng(2017)]{feilee}
L. Fei-Fei and J. Deng.
\newblock Imagenet: Where have we been? where are we going?
\newblock In \emph{CVPR Beyond ImageNet Large Scale Visual Recognition Challenge Workshop}, 2017.

\bibitem[Gutmann and Hyv\"{a}rinen(2012)]{infoNCE_2012}
Michael~U. Gutmann and Aapo Hyv\"{a}rinen.
\newblock Noise-contrastive estimation of unnormalized statistical models, with applications to natural image statistics.
\newblock \emph{Journal of Machine Learning Research}, 13:\penalty0 307–361, 2012.

\bibitem[Haykin(2009)]{haykin2009neural}
Simon~S. Haykin.
\newblock \emph{Neural networks and learning machines}.
\newblock Pearson Education, third edition, 2009.

\bibitem[He et~al.(2020)He, Fan, Wu, Xie, and Girshick]{moco_2020}
Kaiming He, Haoqi Fan, Yuxin Wu, Saining Xie, and Ross Girshick.
\newblock Momentum contrast for unsupervised visual representation learning.
\newblock In \emph{2020 IEEE/CVF Conference on Computer Vision and Pattern Recognition (CVPR)}. IEEE, 2020.

\bibitem[Krizhevsky and Hinton(2009)]{cifar10}
Alex Krizhevsky and Geoffrey Hinton.
\newblock Learning multiple layers of features from tiny images.
\newblock Technical report, University of Toronto, 2009.

\bibitem[Le and Yang(2015)]{tinyimagenet}
Ya Le and Xuan~S. Yang.
\newblock Tiny imagenet visual recognition challenge.
\newblock 2015.

\bibitem[Mishra et~al.(2021)Mishra, Shah, Bansal, Jagannatha, Anjaria, Sharma, Jacobs, and Krishnan]{object_aware_2021}
Shlok Mishra, Anshul Shah, Ankan Bansal, Abhyuday Jagannatha, Janit Anjaria, Abhishek Sharma, David Jacobs, and Dilip Krishnan.
\newblock Object-aware cropping for self-supervised learning.
\newblock \emph{arXiv preprint arXiv:2112.00319}, 2021.

\bibitem[Pang et~al.(2022)Pang, Zhang, Li, Cai, and Lu]{pang2022unsupervised}
Bo Pang, Yifan Zhang, Yaoyi Li, Jia Cai, and Cewu Lu.
\newblock Unsupervised visual representation learning by synchronous momentum grouping.
\newblock In \emph{European Conference on Computer Vision}, pages 265--282. Springer, 2022.

\bibitem[Peng et~al.(2022)Peng, Wang, Zhu, Wang, and You]{Peng_2022}
Xiangyu Peng, Kai Wang, Zheng Zhu, Mang Wang, and Yang You.
\newblock Crafting better contrastive views for siamese representation learning.
\newblock In \emph{IEEE/CVF Conference on Computer Vision and Pattern Recognition (CVPR)}. IEEE, 2022.

\bibitem[Shorten and Khoshgoftaar(2019)]{shorten2019survey}
Connor Shorten and Taghi~M Khoshgoftaar.
\newblock A survey on image data augmentation for deep learning.
\newblock \emph{Journal of Big Data}, 6\penalty0 (1):\penalty0 1--48, 2019.

\bibitem[Sohn(2016)]{nxtloss}
Kihyuk Sohn.
\newblock Improved deep metric learning with multi-class n-pair loss objective.
\newblock In \emph{Advances in Neural Information Processing Systems}. Curran Associates, Inc., 2016.

\bibitem[Yeh et~al.(2022)Yeh, Hong, Hsu, Liu, Chen, and LeCun]{yeh2022decoupled}
Chun-Hsiao Yeh, Cheng-Yao Hong, Yen-Chi Hsu, Tyng-Luh Liu, Yubei Chen, and Yann LeCun.
\newblock Decoupled contrastive learning.
\newblock In \emph{European Conference on Computer Vision}, pages 668--684. Springer, 2022.

\end{thebibliography}
}


\end{document}